%% file: corl_main.tex
\documentclass{article}

\usepackage[preprint]{corl_2024} 

\usepackage{amssymb}            
\usepackage{mathtools}          
\usepackage{mathrsfs}           
\mathtoolsset{showonlyrefs}     
\usepackage{graphicx}           
\usepackage{subcaption}         
\usepackage[space]{grffile}     
\usepackage{url}                

\usepackage[utf8]{inputenc} 
\usepackage[T1]{fontenc}    
\usepackage{hyperref}       
\usepackage{url}            
\usepackage{booktabs}       
\usepackage{amsfonts}       
\usepackage{nicefrac}       
\usepackage{microtype}      
\usepackage{xcolor}         
\usepackage{graphicx}
\usepackage{amsmath}
\usepackage{bbm}

\usepackage{bm}

\input{macros}

\title{Uniformly Safe RL with Objective Suppression for Multi-Constraint Safety-Critical Applications}

\author{
	Zihan~Zhou\\
	University of Toronto\\
	Vector Institute \\
	\texttt{footoredo@gmail.com}
	\And
	Jonathan~Booher\\
	Nuro Inc.\\
	\texttt{jbooher@nuro.ai}\\
	\And
	Khashayar~Rohanimanesh\\
	Nuro Inc.\\
	\texttt{krohanimanesh@nuro.ai}\\
	\And
	Wei~Liu\\
	Nuro Inc.\\
	\texttt{w@nuro.ai}\\
	\And
	Aleksandr~Petiushko\\
	Nuro Inc.\\
	\texttt{apetiushko@nuro.ai}\\
	\And
	Animesh Garg\\
	Georgia Tech\\
	\texttt{animesh.garg@gatech.edu}
}

\begin{document}
\maketitle

\begin{abstract}
  \input{00abstract}
\end{abstract}


\input{01intro}

\input{02related}

\input{03prelim}

\input{04method}

\input{05expr}

\input{06conclusion}

\bibliography{main}  

\end{document}

%% file: macros.tex
\usepackage{comment,url,algorithm,algorithmic,graphicx,subcaption,relsize}
\usepackage{amssymb,amsfonts,amsmath,amsthm,amscd,dsfont,mathrsfs,mathtools,nicefrac}
\usepackage{float,psfrag,epsfig,color,xcolor,url,hyperref}
\usepackage{epstopdf,bbm,mathtools,enumitem}
\usepackage{stackrel}

\def\balign#1\ealign{\begin{align}#1\end{align}}
\def\baligns#1\ealigns{\begin{align*}#1\end{align*}}
\def\balignat#1\ealign{\begin{alignat}#1\end{alignat}}
\def\balignats#1\ealigns{\begin{alignat*}#1\end{alignat*}}
\def\bitemize#1\eitemize{\begin{itemize}#1\end{itemize}}
\def\benumerate#1\eenumerate{\begin{enumerate}#1\end{enumerate}}

\newenvironment{talign*}
 {\csname align*\endcsname}
 {\endalign}
\newenvironment{talign}
 {\csname align\endcsname}
 {\endalign}

\def\balignst#1\ealignst{\begin{talign*}#1\end{talign*}}
\def\balignt#1\ealignt{\begin{talign}#1\end{talign}}

\let\originalleft\left
\let\originalright\right
\renewcommand{\left}{\mathopen{}\mathclose\bgroup\originalleft}
\renewcommand{\right}{\aftergroup\egroup\originalright}

\def\tinycitep*#1{{\tiny\citep*{#1}}}
\def\tinycitealt*#1{{\tiny\citealt*{#1}}}
\def\tinycite*#1{{\tiny\cite*{#1}}}
\def\smallcitep*#1{{\scriptsize\citep*{#1}}}
\def\smallcitealt*#1{{\scriptsize\citealt*{#1}}}
\def\smallcite*#1{{\scriptsize\cite*{#1}}}

\def\mbb#1{\mathbb{#1}}

\def\reals{\mathbb{R}} 

\def\<{\left\langle} 
\def\>{\right\rangle}

\def\E{\mbb{E}} 

\def\Esubarg#1#2{\E_{#1}\left[{#2}\right]}

\newcommand{\grad}{\nabla}

\ifdefined\nonewproofenvironments\else
\ifdefined\ispres\else

\newenvironment{proof-sketch}{\noindent\textbf{Proof Sketch}
  \hspace*{1em}}{\qed\bigskip\\}
\newenvironment{proof-idea}{\noindent\textbf{Proof Idea}
  \hspace*{1em}}{\qed\bigskip\\}
\newenvironment{proof-of-lemma}[1][{}]{\noindent\textbf{Proof of Lemma {#1}}
  \hspace*{1em}}{\qed\\}
\newenvironment{proof-of-theorem}[1][{}]{\noindent\textbf{Proof of Theorem {#1}}
  \hspace*{1em}}{\qed\\}
\newenvironment{proof-attempt}{\noindent\textbf{Proof Attempt}
  \hspace*{1em}}{\qed\bigskip\\}

\fi

\fi

\hypersetup{
  colorlinks,
  linkcolor={red!50!black},
  citecolor={blue!50!black},
  urlcolor={blue!80!black}
}

\newcommand{\ind}[1]{{\mathbbm{1}}_{\{ #1 \}} }

\newcommand{\handout}[5]{
  \noindent
  \begin{center}
    \framebox{
      \vbox{
        \hbox to 5.78in { {\bf \title } \hfill #2 }
        \vspace{4mm}
        \hbox to 5.78in { {\Large \hfill #5  \hfill} }
        \vspace{2mm}
        \hbox to 5.78in { {\em #3 \hfill #4} }
      }
    }
  \end{center}
  \vspace*{4mm}
}

\newcommand{\states}{\mathcal{S}}
\newcommand{\actions}{\mathcal{A}}
\newcommand{\constraints}{\mathcal{C}}

\newcommand{\cgamma}[1]{\gamma_{C_{#1}}}

\newcommand{\RV}{V_R}
\newcommand{\RQ}{Q_R}

\newcommand{\RJ}{\mathcal{J}_R}
\newcommand{\CV}[1]{V_{C_{#1}}}
\newcommand{\CQ}[1]{Q_{C_{#1}}}

\newcommand{\CJ}[1]{\mathcal{J}_{C_{#1}}}
\newcommand{\SWJ}{\mathcal{J}_{switch}}
\newcommand{\SPJ}{\mathcal{J}_{supp}}

\newcommand{\vind}[2]{z_{#1}^{#2}}
\newcommand{\nvind}[1]{\vind{#1}{-}}

%% file: 00abstract.tex

Safe reinforcement learning tasks are a challenging domain despite being very common in the real world. The widely adopted CMDP model constrains the risks in expectation, which makes room for dangerous behaviors in long-tail states. In safety-critical domains, such behaviors could lead to disastrous outcomes. To address this issue, we first describe the problem with a stronger \textbf{Uniformly Constrained MDP (UCMDP)} model where we impose constraints on all reachable states; we then propose \textbf{Objective Suppression}, a novel method that adaptively suppresses the task reward maximizing objectives according to a safety critic, as a solution to the Lagrangian dual of a UCMDP. We benchmark Objective Suppression in two multi-constraint safety domains, including an autonomous driving domain where any incorrect behavior can lead to disastrous consequences. On the driving domain, we evaluate on open source and proprietary data and evaluate transfer to a real autonomous fleet. Empirically, we demonstrate that our proposed method, when combined with existing safe RL algorithms, can match the task reward achieved by baselines with significantly fewer constraint violations.

%% file: 01intro.tex
\section{Introduction}

Reinforcement learning (RL) is a general approach to solving challenging tasks in many domains such as robotics, navigation, and even generative modeling~\cite{Mnih2013PlayingAW, Silver2017MasteringTG, Gu2016DeepRL, Kiran2020DeepRL}. Policies learned with RL seek to maximize a ``task reward'' over a long run which can be specified either via manually defined functions or through learned models. However, without careful tuning of this reward function, it can be difficult for policies to learn to perform well in safety-critical situations like those in the autonomous driving domain~\cite{Amodei2016ConcretePI, Kiran2020DeepRL}. Safe RL~\cite{Gu2022ARO} attempts to solve this by modeling RL with safety concerns as a constrained optimization problem and has been widely adopted in safety-critical RL problems~\cite{Cheng2019EndtoEndSR, ShalevShwartz2016SafeMR}

One common practice of safe RL is to use the Constrained Markov Decision Process (CMDP, \cite{Altman1999ConstrainedMD, Achiam2017ConstrainedPO}) model. CMDPs characterize safety concerns as state-action-dependent scalar constraint functions, similar to the task reward. The safety guarantee is then realized by constraining the expected return of the constraint functions expressed as \cite{ShalevShwartz2016SafeMR}:
\begin{equation}
C_i : J_{C_i}(\pi) = \mathbb{E}_{\tau \sim \pi} \left[ \sum_{t=0}^{\infty} \gamma^t C_i(s_t, a_t, s_{t+1})\right]
\label{eq:cpo}
\end{equation}
 However, one drawback of this formulation is that constraining only the expected return might lead to high-risk behaviors in less frequently visited states according to $\pi$, which can lead to disastrous outcomes in real-world safety-critical domains. Addressing this issue is particularly important in domains such as self-driving cars, where it is crucial not to compromise critical safety constraints (e.g., prioritizing human safety above all) for the sake of achieving task-level objectives, even in rare long-tail catastrophic events.
 
 We thus propose {\em Uniformly Constrained MDP (UCMDP)}, a variant of CMDP which removes the expectation imposed by the task policy $\pi$ in the constraints objective in  Equation~\ref{eq:cpo} by imposing constraints uniformly across all reachable states. This reformulation effectively tightens the safety constraints to ensure optimization toward worst-case scenarios and rare long-tail events. To solve the UCMDP, we use primal-dual methods \cite{JMLR:v18:15-636}, transforming UCMDP into its Lagrangian dual form, with state-dependent Lagrangian multipliers.

We then derive a non-parametric approach that approximates the state-dependent Lagrangian multipliers from objective-switching methods~\cite{zhou2022continuously}. The resulting algorithm is our proposed \textbf{Objective Suppression}, a safe RL algorithm that makes adaptive choices of suppressing and balancing the task reward maximizing objective and constraint-satisfying objectives, as a solution to the UCMDP model. Furthermore, the primal-dual nature of Objective Suppression allows us to organically combine it with existing hierarchical safe-RL methods such as Recovery RL~\cite{Thananjeyan2020RecoveryRS}. 

In multi-constraint safe RL domains, methods that rely on linearly combining the task reward and constraints can struggle to assign a set of weights for all the constraints without some of the constraints being overshadowed by others; hierarchical methods, on the other hand, can face difficulties in building multiple hierarchies. We then proceed to show that by combining Objective Suppression with hierarchical methods, we can utilize both regimes to accommodate different sources of safety constraint.

We empirically test our method in two challenging domains featuring multiple constraints: a \emph{Mujoco-Ant}~\citep{todorov2012mujoco, 1606.01540} domain with dynamic obstacles, where our method lowers the number of collisions by 33\%; and the \emph{Safe Bench}~\citep{safebench} domain, where our method reduces the constraint violations by at least half. In both domains, our method achieves the results without significant sacrifice of task return. We also discuss our experiments conducted on both open-source and proprietary data, and evaluate their transfer to a real autonomous fleet.


We summarize our core contributions as follows:

\begin{itemize}
    \item We propose \textbf{UCMDP}, a variant of CMDP that puts uniform attention on all states to prevent dangerous behaviors in long-tail cases.
    \item We propose \textbf{Objective Suppresion}, a practical algorithm that solves UCMDPs that can be applied to existing safe RL methods.
    \item We adapt two \textbf{multi-constraint} safe RL benchmarks with continuous state and action space from Mujoco-Ant and Safe Bench. We then empirically show that Objective Suppresion significantly improves safety in the proposed benchmarks.
\end{itemize}

%% file: 02related.tex
\section{Related Works}


\paragraph{Lagrangian Methods for CMDPs} Lagrangian relaxation~\citep{Altman1999ConstrainedMD, bertsekas2016nonlinear} uses a primal-dual method to turn the constrained optimization problem of safe RL into an unconstrained one. \citet{JMLR:v18:15-636} demonstrates Lagrangian relaxation can be adopted to safe RL problems and achieve satisfying empirical performances. RCPO~\citep{DBLP:conf/iclr/TesslerMM19} turns Lagrangian multipliers into reward penalty weights to reach constrained goals with both theoretical and empirical evidence. \citet{Stooke2020ResponsiveSI} takes Lagrangian methods into controls domain by utilizing the derivatives of the constraint function. Objective Suppression solves an approximation of the Lagrangian dual of UCMDP.

\paragraph{Hierarchical Methods for CMDPs} Another way of solving safe RL problems is to apply a safety layer~\citep{Dalal2018SafeEI} on top of the task reward-maximizing policy. \citet{Dalal2018SafeEI} derives a closed-form solution for action correction by learning a linearized model. Recovery RL~\citep{Thananjeyan2020RecoveryRS} learns a parameterized recovery policy that leads the agent away from dangerous areas. \citet{bharadhwaj2021conservative} takes inspiration from offline RL algorithms and proposes a hierarchical method with a dynamic error tolerance threshold that achieves provable per-iteration safety guarantees. 


\paragraph{Learning in Autonomous Vehicles} Most prior work in learning-based methods in AV relies on behavior cloning approaches~\citep{uniad,reasonnet}. Some methods add layers of hierarchy~\citep{mgail} in order to learn improved BC policies but can still struggle with safety. Pure cloning-based approaches have been demonstrated to be insufficient to handle long-tail cases or ensure safety~\citep{imitation-not-enough,waymax} which motivates the exploration of RL-based approaches. 


%% file: 03prelim.tex
\section{Preliminaries}

\paragraph{Constrained MDPs} We describe a safe RL problem under the assumption of a constrained Markov Decision Process (CMDP, \cite{Altman1999ConstrainedMD, Garca2015ACS, Achiam2017ConstrainedPO}), where the agent is required to maximize its expected return while ensuring all the safety constraints are met. Formally, a CMDP is defined as a 7-tuple $(\states, \actions, P, R, \gamma, \constraints, \epsilon,\mu_0)$, where $\states$ is the state space, $\actions$ is the action space, $P:\states\times\actions\times\states\rightarrow [0,1]$ is the stochastic state transition function, $R:\states\times\actions\times\states\rightarrow\reals$ is the task reward function, $\gamma$ is the discount factor, $\constraints=\{(C_i:\states\rightarrow\{0,1\}, \cgamma{i})\}$ is the set of constraints and their corresponding discount factors, $\epsilon\in\reals^+$ is the constraint violation limit, and $\mu_0:\states\rightarrow [0,1]$ is the initial state distribution. Let $n=|\constraints|$ denote the number of constraints. In the context of safe RL, these constraints are also \emph{risks}.

For a stochastic policy $\pi:\states\times\actions\rightarrow [0,1]$, the task state value function and state-action value function are:

\begin{align}
    \RV^\pi(s)&=\Esubarg{\tau\sim \pi}{\sum_{t=0}^\infty \gamma^t R(s_t,a_t,s_{t+1})|s_0=s}\\
    \RQ^\pi(s, a)&=\Esubarg{\tau\sim \pi}{\sum_{t=0}^\infty \gamma^t R(s_t,a_t,s_{t+1})|s_0=s,a_0=a}\\
\end{align}

Similarly, the state value function and state-action value function for risk constraints $i\in\{1,\dots n\}$ are:

\begin{align}
    \CV{i}^\pi(s)&=\Esubarg{\tau\sim \pi}{\sum_{t=0}^\infty \cgamma{i}^t C_i(s_t)|s_0=s}\\
    \CQ{i}^\pi(s, a)&=\Esubarg{\tau\sim \pi}{\sum_{t=0}^\infty \cgamma{i}^t C_i(s_t)|s_0=s,a_0=a}\\
\end{align}


Let $\RJ^\pi=\Esubarg{s\sim\mu_0}{\RV^\pi(s)}$ and $\CJ{i}^\pi=\Esubarg{s\sim\mu_0}{\CV{i}^\pi(s)}$. The objective of a safe RL problem is to solve the following constrained optimization problem:

\begin{equation}
    \pi^*=\arg\max_\pi \RJ^\pi\quad\text{s.t.}\ \CJ{i}^\pi\leq\epsilon\quad i\in\{1,\dots, n\}\label{eq:cmdp}
\end{equation}




\paragraph{Lagrangian methods} A classic way of solving constrained optimization problems is Lagrangian methods. For a real vector $\mathbf{x}$, the constrained optimization problem formulated by

\begin{equation}
    \max_{\mathbf{x}}f(\mathbf{x})\quad\text{s.t. } g_i(\mathbf{x})\leq 0\quad i\in\{1,\dots,m\} \label{eq:lag}
\end{equation}

can be solved by introducing a Lagrangian dual variable $\boldsymbol{\lambda}\in\reals_+^m$ and the Lagrangian function $\mathcal{L(\mathbf{x}, \boldsymbol{\lambda})}=f(\mathbf{x})-\boldsymbol{\lambda}^\top \mathbf{g}(\mathbf{x})$. The optimal solution to the dual problem $\min_{\boldsymbol{\lambda}\geq 0}\max_\mathbf{x} \mathcal{L(\mathbf{x}, \boldsymbol{\lambda})}$ is an upper bound of the original constrained problem, which is typically solved with iterative gradient descent:

\begin{align}
    &\grad_{\mathbf{x}}\mathcal{L}(\mathbf{x},\boldsymbol{\lambda})=\grad_{\mathbf{x}}f(\mathbf{x})-\boldsymbol{\lambda}^{\top} \grad_{\mathbf{x}}\mathbf{g}(\mathbf{x})\\
    -& \grad_{\boldsymbol{\lambda}}\mathcal{L}(\mathbf{x},\boldsymbol{\lambda})= \mathbf{g}(\mathbf{x})
\end{align}

\paragraph{Hierarchical methods} One regime of solving CMDPs is to apply a safety layer to adjust the unsafe actions~\cite{Dalal2018SafeEI, Thananjeyan2020RecoveryRS, Yu2022SEditor}. The safety layer overwrites the proposed actions that are deemed unsafe according to the safety critic $\CQ{i}^\pi$:

\begin{equation}
    a_t=\left\{
\begin{array}{ll}
    a\sim\pi(s) & \text{if $\CQ{i}^\pi(s_t,a)\leq \epsilon$ for all $i\in\{1,\dots,n\}$,} \\
    a\sim\sigma(s) & \text{otherwise.}
\end{array}
    \right.
\end{equation}

$\sigma:\states\times\actions\rightarrow [0,1]$ is a safety-ensuring policy. One choice of $\sigma$ is to project the actions into a feasible space~\citep{Pham2017OptLayerP, Dalal2018SafeEI}, e.g., $\sigma(s)=\arg\min_a\|a-\tilde{a}\|\ \text{s.t.}\ \CQ{i}^\pi(s_t, a)\leq \epsilon\ \text{for all}\ i\in\{1,\dots,n\}$, where $\tilde{a}\sim\pi(s)$. Another choice is to use a separate parameterized policy~\citep{Thananjeyan2020RecoveryRS, Yu2022SEditor}. Recovery RL~\citep{Thananjeyan2020RecoveryRS} trains a recovery policy to minimize the constraint violations, i.e., $\sigma^*=\arg\min_\sigma\sum_{i=1}^n w_i\CJ{i}^{(\pi, \sigma)}$.


\paragraph{Policy parameterization} In this work we consider parameterized policies denoted as $\pi_\theta$. The derived gradients~\citep{Williams1992SimpleSG} of $\RJ^{\pi_\theta}$ and $\CJ{i}^{\pi_\theta}$ are:


\begin{equation}
    \grad_{\theta}\RJ^{\pi_\theta}=\Esubarg{\tau\sim\pi_\theta}{\RQ(s_t,a_t)\grad_{\theta}\log\pi_\theta(s_t,a_t)}\label{eq:task-grad},
\end{equation}


\begin{equation}
    \grad_{\theta}\CJ{i}^{\pi_\theta}=\Esubarg{\tau\sim\pi_\theta}{\CQ{i}(s_t,a_t)\grad_{\theta}\log\pi_\theta(s_t,a_t)}\label{eq:risk-grad}.
\end{equation}

Since we always assume using a policy parameterized by $\theta$, we will omit the $\theta$ subscript of $\pi$ onwards for simplicity.

%% file: 04method.tex
\section{Objective Suppression}

We propose a new method to enforce safety constraints on policy optimization based on adaptively suppressing the task reward objectives that combines well with other safe RL algorithms such as Recovery RL in multiple constraint scenarios. We first formulate our safety-critical problem setting with a stronger model than CMDPs, the \textit{Uniformly Constrained MDPs} (UCMDPs) in Section~\ref{sec:ucmdps}. In Section~\ref{sec:algo}, we motivate our proposed \textit{Objective Suppression} method from a reward-switching perspective, as a non-parametric approximation to the solution of the Lagrangian dual of UCMDPs. In Section~\ref{sec:comb}, we discuss the potential of combining \textit{Objective Suppression} with other safe RL algorithms. Finally, in Section~\ref{sec:hrl}, we make a connection from Objective Suppression to Hierarchical RL to provide more insights to our method.

\subsection{Uniformly Constrained MDPs}\label{sec:ucmdps}

CMDPs constraint behaviors in expectation. However, in safety-critical domains such as autonomous driving, safety guarantees in less frequently visited situations should be treated with equal attention. In light of this, we propose a stronger version of CMDPs, named \textbf{Uniformly Constrained MDPs (UCMDPs)}. UCMDPs share the same definition as the CMDPs except for the optimization objective constraint, where we pose a uniform one instead of the expectation one in CMDPs:

\begin{equation}
    \theta^*=\arg\max_\theta \RJ^{\pi}\quad\text{s.t.}\ Q^{\pi}_{C_i}(s,a)\leq\epsilon\quad i\in\{1,\dots,n\}, d^{\pi}(s,a)>0\label{eq:ucmdp}
\end{equation}
$d^{\pi}(s,a)$ is the distribution of state-action pairs under ${\pi}$.

\paragraph{Comparison with SCMDPs} A similar model that focuses on state-wise constraints is \textit{State-wise Constrained MDP} (SCMDP, \citep{2023arXiv230203122Z}). UCMDPs differ from SCMDPs in that UCMDPs constrain the policy-dependent expected return of constraint violations ($Q_{C_i}^\pi$) instead of a purely state-based constraint function. In dynamic environments such as driving, there are often states that are inherently risky due to other actors or environment conditions. As an example, consider a narrow corridor where two vehicles need to negotiate. Entering the corridor is inherently more risky than yielding due to the uncertainty over the behavior of the other agent. In the SCMDP formulation, we would disallow entering the region with a pure state-based constraint function resulting in over conservative exploration. However, in the UCMDP formulation, the policy would be allowed to explore the region and discover a safe negotiation of the narrow corridor. 



\paragraph{Lagrangian dual of UCMDPs} The Lagrangian dual problem of \eqref{eq:ucmdp} is 

\begin{equation}
    \min_{\lambda\geq 0} \max_\theta\mathcal{L}(\theta,\lambda),
\end{equation}
where 
\begin{align}
    \mathcal{L}(\theta,\lambda)= \RJ^{\pi_\theta}-\sum_{i,s,a}\ind{d^{\pi}(s,a)>0}\lambda_i(s,a)\big(Q^{\pi}_{C_i}(s,a)-\epsilon\big),
\end{align}


which we further relax to: 

\begin{align}
    \tilde{\mathcal{L}}(\theta,\lambda)&= \RJ^\pi-\sum_{i,s,a}d^\pi(s,a)\lambda_i(s,a)\big(Q^\pi_{C_i}(s,a)-\epsilon\big)\\
    &= \Esubarg{s,a}{Q^\pi_R(s,a)-\sum_{i=1}^n \lambda_i(s,a)\big(Q^\pi_{C_i}(s,a)-\epsilon\big)}\label{eq:lag-ucmdp}
\end{align}


The gradients for \eqref{eq:lag-ucmdp} are:

\begin{align}
    \grad_{\theta}\tilde{\mathcal{L}}(\theta,\lambda)&=\Esubarg{s,a}{\left(Q^\pi_R(s,a)-\sum_{i=1}^n\lambda_i(s,a)\big(Q^\pi_{C_i}(s,a)-\epsilon\big)\right)\grad_\theta \log \pi(s,a)}\\
     &  =\Esubarg{s,a}{\left(Q^\pi_R(s,a)-\sum_{i=1}^n\lambda_i(s,a)Q^\pi_{C_i}(s,a)\right)\grad_\theta \log \pi(s,a)}\\
     & \qquad +\epsilon\cdot \Esubarg{s,a}{\sum_{i=1}^n\lambda_i(s,a)\grad_\theta \log \pi(s,a)} \label{eq:grad-lag-ucmdp-1}\\
    -\grad_{\lambda}\tilde{\mathcal{L}}(\theta,\lambda)&=\Esubarg{s,a}{\sum_{i=1}^n Q^\pi_{C_i}(s,a)-\epsilon}\label{eq:grad-lag-ucmdp-2}
\end{align}


\eqref{eq:grad-lag-ucmdp-1} is the standard policy gradient with a linear combination of the task objective $Q^\pi_R(s,a)$ and the risk-avoiding objectives $Q^\pi_{C_i}(s,a)$, the latter weighted by the Lagrangian multipliers $\lambda_i(s,a)$. The remaining problem is to maintain the $\lambda_i(s,a)$ with \eqref{eq:grad-lag-ucmdp-2} over the potentially continuous state and action space. One intuitive way is to use a parameterized $\lambda$ function which can be updated with gradient descent. However, introducing a new set of parameters will cause problems such as separate hyperparameter tuning and increased computational cost. In the following section, we instead propose a non-parametric way of approximating $\lambda$.

\subsection{Objective Suppression}\label{sec:algo}


To enforce safety constraints on policy optimization, we would like to train a policy that automatically switches between optimizing for task reward objective $\RJ^\pi$ and risk minimization objective $\CJ{i}^\pi$. One way to accomplish this is to switch the optimization objective in hindsight~\citep{zhou2022continuously}. Specifically, at step $t$ of a trajectory $\tau$, if any risk is encountered after $t$, we then switch to the risk minimization objective; otherwise, we remain using the original task reward objective. Let $\vind{t}{i}(\tau)=\ind{\text{$\tau$ encounters risk $i$ after step $t$}}$ and $\nvind{t}(\tau)=\ind{\text{$\tau$ does not encounter any risk after step $t$}}=\prod_i1-\vind{t}{i}(\tau)$. We define the hard-switching objective as



\begin{equation}
    \grad_{\theta}\SWJ^{\pi}=\Esubarg{\tau}{\left(\nvind{t}(\tau)\RQ^\pi(s_t,a_t)-\sum_{i=1}^nw_i\vind{t}{i}(\tau)\CQ{i}^\pi(s_t,a_t)\right)\grad_{\theta}\log\pi(s_t,a_t)},\label{eq:hard-switch-grad}
\end{equation}


where $w_1,\dots,w_n$ are the weights to balance the different risk minimization objectives. The hard-switching objective \eqref{eq:hard-switch-grad} is an interpolation of $\grad_{\theta}\RJ^{\pi}$ and $\grad_{\theta}\CJ{i}^{\pi}$. However, this objective faces problems in practice. For one, the hard-switching between multiple objectives raises the variance of the gradient, which is widely acknowledged to have a negative impact on training; for another, balancing the different objectives with weights introduces a new set of hyperparameters. If handled improperly, some of the objectives will dominate others and can result in disastrous outcomes. To deal with the aforementioned two problems, we rewrite \eqref{eq:hard-switch-grad} into:

\begin{align}
    \grad_{\theta}\SWJ^{\pi}&=\Esubarg{\tau}{\left(\nvind{t}(\tau)\RQ^\pi(s_t,a_t)-\sum_{i=1}^nw_i\vind{t}{i}(\tau)\CQ{i}^\pi(s_t,a_t)\right)\grad_{\theta}\log\pi(s_t,a_t)}\\
    &=\Esubarg{s_t,a_t}{\left(\mathbin{\color{teal}\Esubarg{\tau}{\nvind{t}(\tau)}}\RQ^\pi(s_t,a_t)-\sum_{i=1}^nw_i\mathbin{\color{purple}\Esubarg{\tau}{\vind{t}{i}(\tau)}}\CQ{i}^\pi(s_t,a_t)\right)\grad_{\theta}\log\pi(s_t,a_t)}\\
    &=\Esubarg{s,a}{\left(\mathbin{\color{teal}p_-(s,a)}\RQ^\pi(s,a)-\sum_{i=1}^nw_i\mathbin{\color{purple}p_i(s,a)}\CQ{i}^\pi(s,a)\right)\grad_{\theta}\log\pi(s,a)}\label{eq:soft-switch-grad}
\end{align}


where $\mathbin{\color{teal}p_-(s,a)}$ is the probability of not encountering any risk after $(s,a)$, and $\mathbin{\color{purple}p_i(s,a)}$ is the probability of encountering risk $i$. 

\paragraph{Practical implementations.} Both the probabilities $p_-(s,a)$ and $p_i(s,a)$ are nontrivial to compute, since we are considering non-halting risks. In practice, we use the state-value function $\tilde{p}_i(s,a)=\CQ{i}^\pi(s,a)$ as a proxy for $p_i(s,a)$. 
For $p_-(s,a)$, notice that $p_-(s,a)\geq\prod_i (1-p_i(s,a))$ and $p_-(s,a)\leq \min_i \{1-p_i(s,a)\}$. We would like a proxy function that is close to 1 when all the risk probabilities are low and close to 0 when at least one of the risk probabilities are high. The specific proxy that we use is $\tilde{p}_-(s,a)=\exp(-\kappa\sum_i\CQ{i}^\pi(s,a))$, with $\kappa$ as the temperature.

We now conclude our Objective Suppression gradient:



\begin{equation}
    \grad_{\theta}\SPJ^{\pi}=\Esubarg{s,a}{\left(\tilde{p}_-(s,a)\RQ^\pi(s,a)-\sum_{i=1}^nw_i\tilde{p}_i(s,a)\CQ{i}^\pi(s,a)\right)\grad_{\theta}\log\pi(s,a)}\label{eq:supp}
\end{equation}


\paragraph{Connection with UCMDPs.} We now show that our Objective Suppression method leads to a solution to the Lagrangian dual problem of the UCMDP. Since in \eqref{eq:supp} we choose $\tilde{p}_-(s,a)=\exp(-\kappa\sum_i\CQ{i}^\pi(s,a))>0$, define $r_i(s,a)=\frac{w_i\tilde{p}_i(s,a)}{\tilde{p}_-(s,a)}$. \eqref{eq:supp} can then be rewritten as

\begin{align}
    \grad_{\theta}\SPJ^{\pi}&=\Esubarg{s,a}{\tilde{p}_-(s,a)\left(\RQ^\pi(s,a)-\sum_{i=1}^nr_i(s,a)\CQ{i}^\pi(s,a)\right)\grad_{\theta}\log\pi(s,a)},
\end{align}

which is a weighted version of \eqref{eq:grad-lag-ucmdp-1}, if we assume small $\epsilon$ and let $\lambda_i\equiv r_i$. By definition, $r_i$ is high when the probability of encountering risk $i$ is high, which balance the Objective Suppression gradient towards avoiding risk $i$, and vice versa when the probability is low. This conceptually serve the similar purpose as the Lagrangian multipliers in the dual problem. 


\subsection{Combining with Existing Safe RL Algorithms}\label{sec:comb}

We empirically find that Objective Suppression works better when combined with other safe RL algorithms. We conjecture that this is because applying multiple regimes of constraint enforcement increases the coverage of different constraints, preventing certain constraints from being dominated by others in one regime. In our experiments, we build our method on top of Recovery RL by enforcing our Objective Suppression objective~\eqref{eq:supp} on both the task and recovery policy.

The original Recovery RL formulation relied on a pre-training stage in order to train the risk critic from demonstrations. This ensures that the method is safe \emph{during exploration}; however, we use a purely online version of Recovery RL where the risk critic is trained jointly with the policy.  

\subsection{Connection to the Hierarchical Reinforcement Learning (HRL) Framework}\label{sec:hrl}

An alternative perspective on our approach is to consider it as a hierarchical reinforcement learning model, specifically within the framework of options \cite{SUTTON1999181}. An option $\mathcal{O} \equiv \langle \mathcal{I}_o, \pi_o, \beta_o \rangle$ defines a temporally extended action (or a policy) where $\mathcal{I}_o(s)$ denotes the set of states where the option can be initiated, $\pi_o(s, a) \mapsto [0, 1]$ denotes the (stochastic) policy associated with that option, and $\beta(s) \mapsto [0, 1]$ denotes the termination condition, or the probability of terminating the option in a given state $s$. 

In our framework, we can view the high level policy $\SWJ^{\pi}$ as a stochastic switching strategy that selects between two options: $\mathcal{O}_{task} \equiv \langle \mathcal{I}_{task}, \pi_R, \beta_{task} \rangle$ and $\mathcal{O}_{recovery} \equiv \langle \mathcal{I}_{recovery}, \pi_C, \beta_{recovery} \rangle$. For option $\mathcal{O}_{task}$, the initiation set $\mathcal{I}_{task}$ denotes the set of risk free states where the agent can safely optimize the task reward, the policy $\pi_R$ denotes the task policy, and the termination condition $\beta_{task}$ gets triggered in any state where any of the constraints (risks) are violated. For option $\mathcal{O}_{recovery}$, the initiation set $\mathcal{I}_{recovery}$ denotes the set of states at which at least one constraint is violated, the policy $\pi_C$ denotes the recovery policy, and the termination condition $\beta_{recovery}$ gets triggered when no constraints (risks) are violated and the agent can safely execute the task policy until the next safety violation. In this view, our framework essentially learns the model of the options $\mathcal{O}_{task}$ and $\mathcal{I}_{recovery}$ in terms of their policies and termination conditions.

%% file: 05expr.tex
\section{Experiments}

\subsection{Environments and baselines}

\paragraph{Safe Mujoco-Ant} This environment is adapted from the 8-DOF \emph{ant-v4} environment from gym~\citep{1606.01540} Mujoco~\citep{todorov2012mujoco}. The main task of the agent is to control the ant to reach a target point. There are two constraints in this environment. The first constraint is to avoid randomly spawned obstacles and the second is to avoid collapsing.

\paragraph{Safe Bench} This environment is a CARLA~\citep{carla} based benchmark for safety in various different driving scenarios~\citep{safebench}. The task reward comes from making progress along the designated route while the constraints come from (1) collisions with obstacles (e.g., pedestrians, curbs, etc) which will terminate the episode, and (2) leaving the lane, i.e. lateral deviation from the lane of travel.
The policy is given access to a bird-eve-view rendering of the scene in addition to a 4-D observation space covering lane placement, speed, and the distances to objects. The action space is a continuous 2-D space consisting of control parameters: acceleration and steering angle. 


\paragraph{Baselines} We test our objective suppression method in the two aforementioned environments, \emph{Safe Mujoco-Ant} and $\emph{Safe Bench}$. We also implement two baselines, a naive \emph{Reward Penalty} baseline, where the optimization objective is a fine-tuned weighted sum of the task reward objective and risk minimization objective; and \emph{Recovery RL}~\citep{Thananjeyan2020RecoveryRS}.

\subsection{Results}

\begin{table}
  \caption{Results in \emph{Safe Mujoco-Ant} collected with 20m environment steps. Average and standard deviation over 5 seeds.}
  \vspace{.5em}
  \label{tab:mujoco-results}
  \centering
  \scalebox{1.0}{
  \begin{tabular}{lllll}
    \toprule
    Name        & Task Reward $\uparrow$ & Collisions $\downarrow$ & Collapse $\downarrow$ \\
    \midrule
    Reward Penalty & 1133.63 $\pm$ 321.17     & 69.70 $\pm$ 32.68           & 0.39 $\pm$ 0.10   \\
    Recovery RL & 1279.62 $\pm$ 143.40     & 6.47 $\pm$ 1.88           & 0.42 $\pm$ 0.20   \\
    Ours ($\kappa=3$)      & 1215.49 $\pm$ 143.40     & 4.36 $\pm$ 0.75           & 0.43 $\pm$ 0.08   \\
    Ours ($\kappa=1$)      & 1175.79 $\pm$ 172.13     & 2.93 $\pm$ 0.67           & 0.54 $\pm$ 0.12   \\
    \bottomrule
  \end{tabular}
  }
\end{table}
\begin{table}
    \caption{Results in \emph{Safe Bench}. All rows given 200k environment steps. Average and standard deviation over 3 seeds.}
  \vspace{.5em}
    \label{tab:safebench-tresults}
    \centering
    \begin{tabular}{llll}
        \toprule
        Name    & Task Reward $\uparrow$    & Collisions $\downarrow$   & Out-of-Lane $\downarrow$ \\
        \midrule
        Reward Penalty & 159.42 $\pm$ 15.96 & 0.55 $\pm$ 0.06 & 63.03 $\pm$ 3.93 \\
        Recovery RL & 124.80 $\pm$ 50.19 & 0.24 $\pm$ 0.16 & 33.35 $\pm$ 19.76 \\
        Ours & 83.17 $\pm$ 23.12 & 0.078 $\pm$ 0.008 & 13.04 $\pm$ 8.82 \\
        \bottomrule
    \end{tabular}
\end{table}

\paragraph{Safe Mujoco-Ant} We compare our method with \emph{Reward Penalty} and \emph{Recovery RL}. Every experiment is run with 5 seeds. The results are shown in Table~\ref{tab:mujoco-results}. For \emph{Reward Penalty}, even with a hyperparameter sweep, we could not find a suitable set of weights to reduce the number of collisions without sacrificing too much task reward. Compared with \emph{Recovery RL}, our method with $\kappa=3$ achieves 33\% less collisions at a mere expense of 5\% less task reward.

We notice from our experimentation that although Recovery RL is able to effectively lower the collisions compared with the Reward Penalty baseline thanks to its policy switching mechanism, it struggles in training a good recovery policy. Even though policy switching eliminates the need to optimize for both task reward and constraint avoidance, in our setting, the two constraint-minimization objectives still constitute a conflicting training objective for the recovery policy, which is further exacerbated by the lack of on-policy examples. In fact, one extremely sensitive hyperparameter to tune for Recovery RL is the weight of the collapse-avoiding objective for the recovery policy. With a high weight, the recovery policy ignores the collision-avoiding objective, resulting in a surge in collisions; with a low weight, the recovery policy tends to collapse in front of obstacles, resulting in a surge in collapsed finishes. The introduction of Objective Suppression effectively alleviates the problem, striking a balance between the two constraints for the recovery policy. This demonstrates how our method can shine in multi-constraint scenarios.

\paragraph{Safe Bench} We compare our method with \emph{Reward Penalty} and \emph{Recovery RL}. Each method is evaluated using 50 rollouts from the policy in different environments. The results are shown in Table~\ref{tab:safebench-tresults}. Compared with the baseline \emph{Recovery RL}, our method outperforms on constraint satisfaction while only incurring a small decrease in task reward. Compared with Recovery RL, our method reduces collision to 32.5\% and out-of-lane violations to 39.1\%.

\paragraph{Proprietary Data and Real World} We also demonstrate a variant of this approach on extensively collected proprietary driving datasets containing thousands of real-world miles covering a variety of challenging conditions. On these datsets, we observe an increase in distance traveled (a proxy for success) combined with a $2\%$ improvement in collision rate, and a $38\%$ improvement in harsh braking events compared to Recovery RL alone. This evaluation in simulation enabled us to deploy our method on a fleet of vehicles in the real world with performance mirroring the results in simulation.


%% file: 06conclusion.tex
\section{Conclusion}

We identified that the widely-used CMDP model for safe RL can potentially allow dangerous behaviors in less frequently visited states. We proposed UCMDPs that uniformly constrain reachable states to counter this issue. To solve UCMDPs, we propose Objective Suppression, a novel algorithm that adaptively suppresses the task reward objective to enforce safety constraints. Combined with existing safe RL algorithms, we demonstrate that Objective Suppression can maintain the task reward of the base algorithm while significantly lowering the constraint violations in multi-constraint scenarios, including an autonomous driving domain where incorrect behaviors can be disastrous.

\section{Limitations}

In our evaluations, we observe that $\kappa$ is still an important tunable hyperparameter for Objective Suppression. It would be interesting to conduct a more thorough study on how this parameter impacts the task reward-risk trade-off. On the other hand, both our benchmarks are continuous control domains, despite our method also applies to discrete action spaces. We believe evaluating Objective Suppression in more domains with discrete action spaces remains an interesting future direction.